\documentclass[letterpaper, 10 pt, conference]{ieeeconf}  
\usepackage{graphicx}
\usepackage{xcolor}

\usepackage{amsmath}

\IEEEoverridecommandlockouts                              

\title{\LARGE \bf
Effective self-righting strategies for elongate multi-legged robots
}

\author{Erik Teder$^{1,*}$, Baxi Chong$^{2,*}$, Juntao He$^{2}$, Tianyu Wang$^{2}$, Massimiliano Iaschi$^{2}$, Daniel Soto$^{2}$, Daniel I Goldman$^{2}$%
\thanks{*These authors contributed equally to this work}
\thanks{$^{1}$Erik Teder is with Hillsdale College, Hillsdale, MI 49242, USA.
        {\tt\small eteder@hillsdale.edu}}%
\thanks{$^{2}$Baxi Chong, Juntao He, Tianyu Wang, Massimiliano Iaschi, Daniel Soto, and Daniel I Goldman are with the Georgia Institute of Technology, Atlanta, GA 30332.
        {\tt\small \{bchong9, jhe391, tianyuwang, miaschi3, dsoto7\}@gatech.edu, {\tt\small daniel.goldman@physics.gatech.edu}}}%
}

\begin{document}

\maketitle
\thispagestyle{empty}
\pagestyle{empty}

\begin{abstract}

Centipede-like robots offer an effective and robust solution to navigation over complex terrain with minimal sensing. However, when climbing over obstacles, such multi-legged robots often elevate their center-of-mass into unstable configurations, where even moderate terrain uncertainty can cause tipping over. Robust mechanisms for such elongate multi-legged robots to self-right remain unstudied.vHerpproach to investigatebiological and robophysical e self-righting strategies. We first released \textit{S. polymorpha} upside down from a 10 cm height and recorded their self-righting behaviors using top and side view high-speed cameras. Using kinematic analysis, we hypothesize that these behaviors can be prescribed by two traveling waves superimposed in the body’s lateral and vertical planes, respectively. We tested our hypothesis on an elongate robot with static (non-actuated) limbs, and we successfully reconstructed these self-righting behaviors. We further evaluated how wave parameters affect self-righting effectiveness. We identified two key wave parameters: the spatial frequency, which characterizes the sequence of body-rolling, and the wave amplitude, which characterizes body curvature. By empirically obtaining a behavior diagram of spatial frequency and amplitude, we identify effective and versatile self-righting strategies for general elongate multi-legged robots, which greatly enhances these robots' mobility and robustness in practical applications such as agricultural terrain inspection and search-and-rescue. 

\end{abstract}

\section{INTRODUCTION}

\begin{figure}
    \centering
    \includegraphics[width=\linewidth]{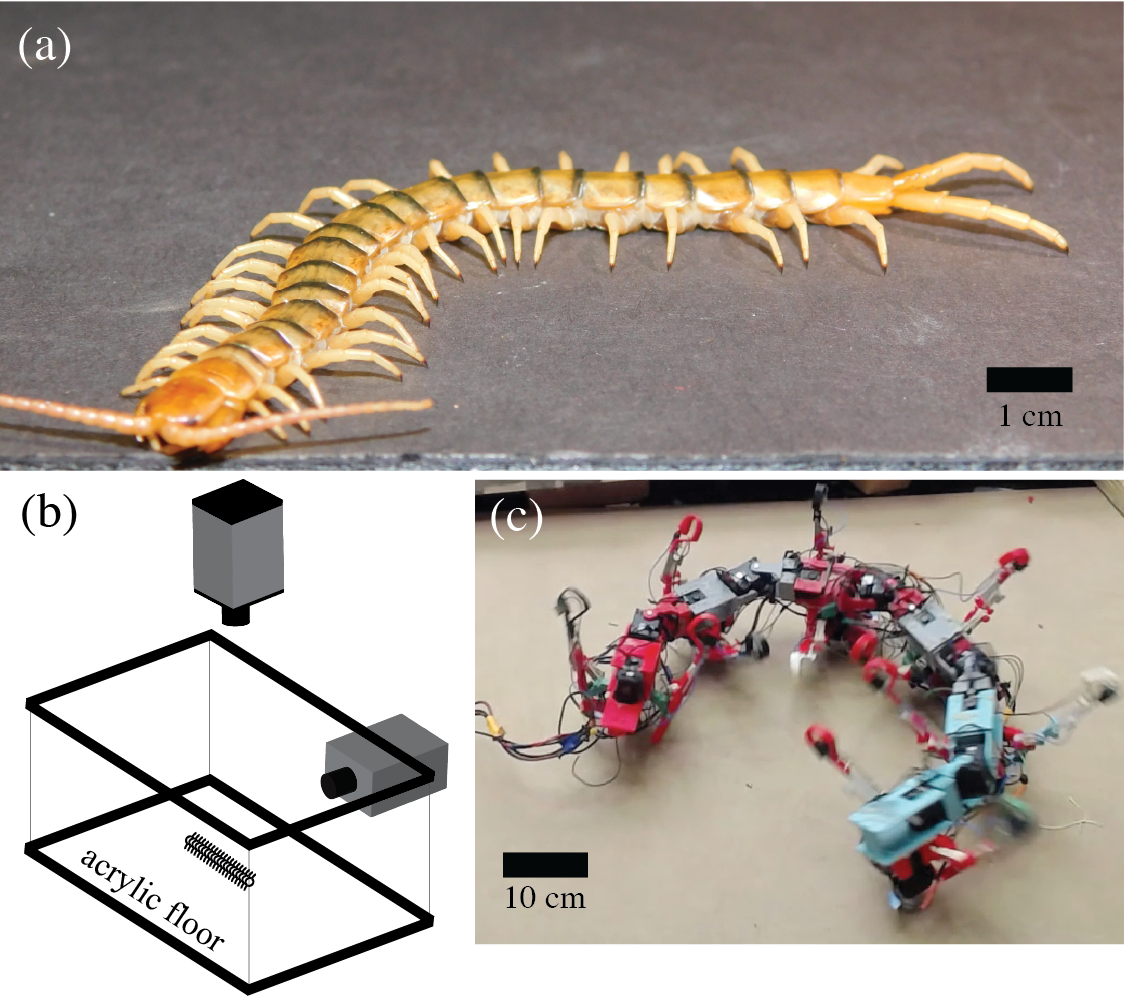}
    \vspace*{-5mm}
    \caption{\textbf{Centipede, centipede robot, and experimental setup.} (a) An adult \textit{S. polymorpha} centipede used in the experiments. (b) The biological experimental setup: self-righting in a glass tank with an acrylic floor is recorded by top and side view high speed cameras. (c) An inverted full-size multi-legged robot necessitating self-righting.}
    \label{fig:fig1}
\end{figure}%

Recently, centipede-like robots, characterized by their elongate segmented bodies and many relatively simple legs, have emerged as an alternative solution for locomotion over complex terrain that does not require complicated sensing and control schemes~\cite{chongScience,ozkansystematic,chong2023self,he2024learning2}, and they have a variety of applications in search-and-rescue \cite{itoRubbles}, agriculture~\cite{pederson2006agriculturalrobots}, and other sectors where low-profile, robust robots are needed. Centipede robots locomote by propagating waves of body undulation and limb stepping, and while this method is effective at rugged terrain traversal, it can elevate the robot's center-of-mass to unstable configurations where the robot can be easily tipped over after obstacle interaction, as in Fig.~\ref{fig:fig1}c. Tipping is best avoided altogether~\cite{roanValidation}, but when it cannot be, a robust strategy for self-righting is needed to ensure reliable locomotion.

A variety of self-righting mechanisms have been developed for robots. For example, RHex, a hexapod with c-shaped rotating limbs, performs aerobatic backflips to self-right \cite{saranli2004rHex}. Jumping/hopping robots use rounded shapes and a low center-of-mass to passively roll back to an upright position~\cite{kovac2009passiveJumper,beyer2009hopping}. Other self-righting strategies employed by robots include using levers, arms, or feet to push themselves upright~\cite{kessens20212framework,peng6MultileggedRighting,yimJumper}. Finally, cockroach-inspired robots use a combination of active wings and a rounded shell to self-right~\cite{liCockroach}. While these methods have been proven successful for their respective morphologies, they cannot be readily generalized to an elongate multi-legged robot because of this robot's drastically different morphology: (1) elongate robots are too heavy to perform aerobatics, (2) their limbs often lack the range of motion needed to push the robot upright, and (3) when inverted, these robots' center-of-mass is often close to the ground, making self-righting challenging. We thus identified the need for an effective self-righting strategy that could be generalized across this different family of robot morphologies. In this paper, we present a comparative approach to investigating self-righting strategies in elongate, multi-legged robots.


We first explore self-righting behaviors in centipedes (Fig.~\ref{fig:fig1}) because they are elongate limbed organisms exhibiting remarkable locomotion performance in complex environments (Sec. II).  We identify two self-righting strategies from observations in centipedes, and we hypothesize that these strategies can be modeled as the superposition of lateral and vertical travelling waves propagated along the body (Sec. III). We reconstruct both strategies using a robophysical model (see Fig.~\ref{fig:fig3}) with static (non-actuated) limbs. We further evaluate how self-righting effectiveness is affected by control parameters and robot morphology (Sec. III). Finally, we observe that  unsuccessful self-righting can result in effective lateral displacement, a phenomenon known as sidewinding in limbless robots/animals (Sec. IV). 

\begin{figure}
    \centering
    \includegraphics[width=\linewidth]{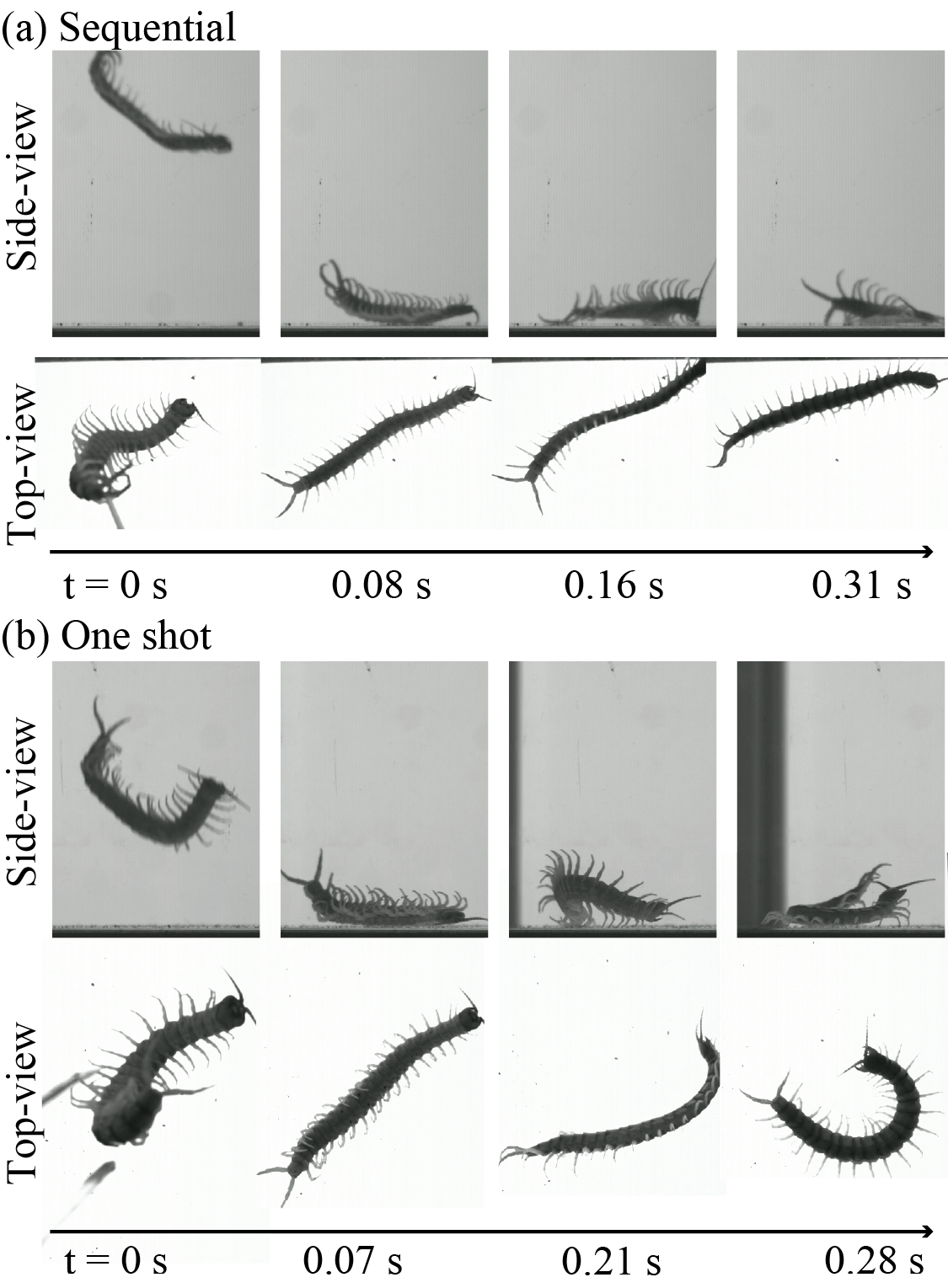}
    \vspace*{-5mm}
    \caption{\textbf{Sequential and One-shot Self-righting Snapshots} (a) Top and side view snapshots of the centipede falling and self-righting using the sequential strategy. The four snapshots indicate the releasing, upside down, initiating self-righting, and successful self-righting respectively. (b) Top and side view of the centipede falling onto its back and self-righting using the one-shot strategy. }
    \label{fig:fig2}
\end{figure}%

\section{SELF-RIGHTING IN CENTIPEDES}

Self-righting is an essential ability for many terrestrial animals, as inverted animals face increased danger from predation and exposure~\cite{rubin2018turtles}. Self-righting has been studied in animals like cockroaches, which use diverse strategies of rolling and pitching with legs and wings~\cite{li2019cockroach}, turtles, which use a variety of strategies thought to be primarily dependent on shell morphology~\cite{rubin2018turtles}, and beetles, which exhibit approximately 20 different self-righting behaviors depending on the species~\cite{frantsevich2004beetles}. However, a research gap exists regarding self-righting in elongate limbed animals like centipedes. Centipedes, composed of many limbed segments forming a flexible body (see Fig.~\ref{fig:fig1}a), possess a drastically different morphology than the rigid animals that have been studied so far, and thus their self-righting strategies stand apart from the others.

We studied self-righting in adult \textit{S. polymorpha} centipedes (body length = 11.3 $\pm$ 0.9 cm) using the experimental apparatus shown in Fig.~\ref{fig:fig1}b, where a glass tank with a hard, smooth acrylic floor was lit from the top and side with LED panels. Attached to the apparatus were top and side view AOS high-speed cameras to capture self-righting at 500 frames per second. We released the centipedes from a 10 cm height onto their backs in the tank and recorded their self-righting behavior using the cameras. The height of 10 cm was chosen because it will not allow sufficient time to perform self-righting midair before landing.

Qualitative kinematic analysis of the resulting videos (3 individuals, 5 trials in each individual) revealed that \textit{S. polymorpha} uses two primary self-righting strategies, which we will refer to as the ``sequential" and ``one-shot" strategies. In the sequential strategy, shown in Fig.~\ref{fig:fig2}a, the centipede starts the self-righting roll from one end of the body and propagates the roll along the body until the animal is upright. In the one-shot strategy, shown in \ref{fig:fig2}b, the animal curls up into a ``C" shape and then flips over all at once as all body segments facilitate self-righting simultaneously. Additionally, we did not observe the active use of limbs during the centipede self-righting. Instead, self-righting was accomplished primarily by body movement.

Quantitative kinematic analysis would present substantial challenges, including tracking the body/leg positions using two cameras. Further, we notice a substantial amount of body/leg intersection in the videos, which further complicates the tracking. While we cannot yet quantify the body kinematics of centipede self-righting (the subject for future work), our kinematic analysis is adequate to construct approximations of these behaviors using a robophysical model.

\section{A MATHEMATICAL MODEL}

\begin{figure}
    \centering
    \includegraphics[width=\linewidth]{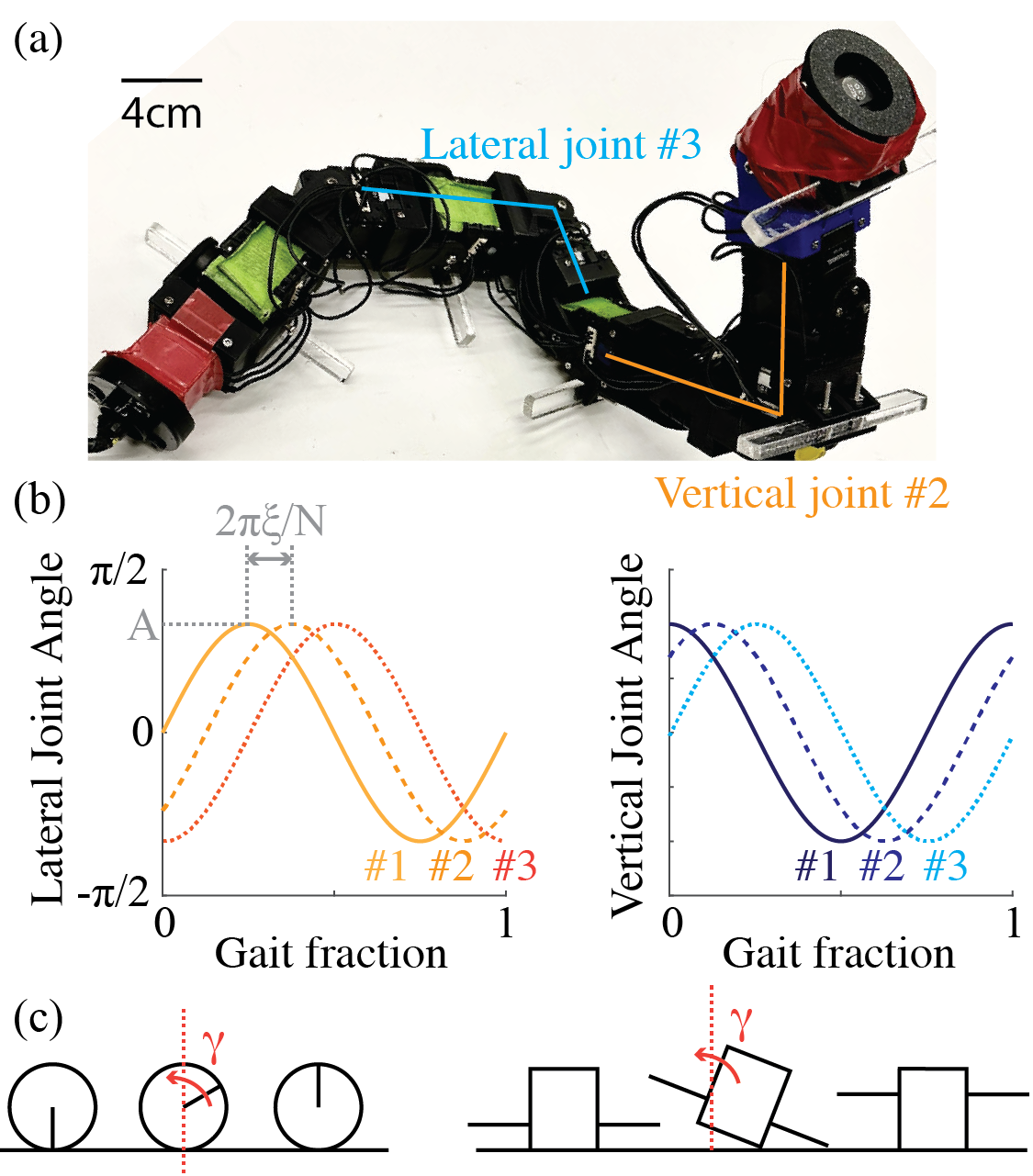}
    \caption{\textbf{Self-righting Gaits} (a) The model is composed of a chain of alternating lateral and vertical servomotors forming lateral and vertical joints, two of which are labelled here. (b) The joint angle prescription of (\textit{left}) lateral and (\textit{right}) vertical joints as a function of gait fraction. Color indicates the joint index. The amplitude of the joint angle is denoted by $A$. The behavior lag between adjacent lateral joints (and identical to the phase lag between adjacent vertical joints) is denoted by $2\pi \xi$/N. (c) Definition of rolling $\gamma$ in (\textit{left}) limbless and (\textit{right}) legged systems.} 
    \label{fig:fig4}
\end{figure}%

Inspired by our observations in biological experiments, we are interested in developing a body-driven self-righting scheme. In limbless literature, rolling gaits are documented and adopted as a common locomotion strategy~\cite{hatton2010generating}. Specifically, consider a robot with alternating pitch and yaw joints (e.g.,~Fig.~\ref{fig:fig4}). Let $\alpha_{l}(i,t)$ and $\alpha_{v}(i,t)$ be the $i$-th yaw (lateral) and pitch (vertical) joint respectively. The rolling gaits are then prescribed as:

\begin{align}\label{eq:roll}
    \alpha_{l}(t,i) = A\sin{\omega t} \nonumber \\
    \alpha_{v}(t,i) = A\cos{\omega t} 
\end{align}

\noindent where $A$ and $\omega$ are the amplitude and temporal frequency of the rolling respectively.  Note that all lateral/vertical joints share the identical joint angle. During rolling gaits, the locomotor experiences a periodic transition from upside down to upside right (self-righting), and then returns to the upside down position (``self-wronging"). We model the one-shot self-righting as one half of the rolling gait: $\omega t\in [0\ \pi]$. 

We next explore how our gait equation can be extended to sequential self-righting behaviors. Sidewinding is another commonly used locomotion strategy for limbless robots that includes body undulations in both the lateral and vertical planes~\cite{astley2015modulation,chong2022general,kojouharov2024anisotropic}. Further, a substantial amount of rolling is observed if the vertical/lateral body undulation is not properly synchronized~\cite{zhong2020frequency,chong2022general}. Specifically, the sidewinding gait equations are prescribed as:

\begin{align}\label{eq:sidewind}
    \alpha_{l}(t,i) = A\sin{(\omega t + 2\pi \xi \frac{i}{N})} \nonumber \\
    \alpha_{v}(t,i) = A\cos{(\omega t + 2\pi \xi \frac{i}{N})}
\end{align}

\noindent where $N$ is the number of lateral joints and $\xi$ is the spatial frequency. Notably, Eq.~\ref{eq:roll} is a special case of Eq.~\ref{eq:sidewind} with $\xi=0$. In other words, rolling gaits are one special example of sidewinding gaits.

In our prior work on sidewinding, we increased the spatial frequency ($\xi$) to increase static stability by minimizing rolling~\cite{chong2021frequency}. Here, with an opposite goal of maximizing rolling, we modulate the frequency by decreasing~$\xi$. 

We quantify the rolling displacement by $\gamma$ (Fig.~3.c). $\Delta \gamma$ characterizes the displacement in the rolling direction (in the transverse plane). For an ideal limbless robot, we have $\Delta \gamma = \omega t$, where a complete gait cycle indicates a complete roll over. 

We then consider irregular (e.g. limbed) shapes (e.g., Fig.~3.c. right panel). In those cases, there are only two stable configurations: $\gamma = 0$ or $\gamma = \pi$. The energy barrier between those two configurations is then determined by the leg size. In this way, the presence of legs can introduce a non-linear relationship between $\Delta\gamma$ and $\omega t$.

\section{A ROBOPHYSICAL MODEL}

\begin{figure}
    \centering
    \includegraphics[width=\linewidth]{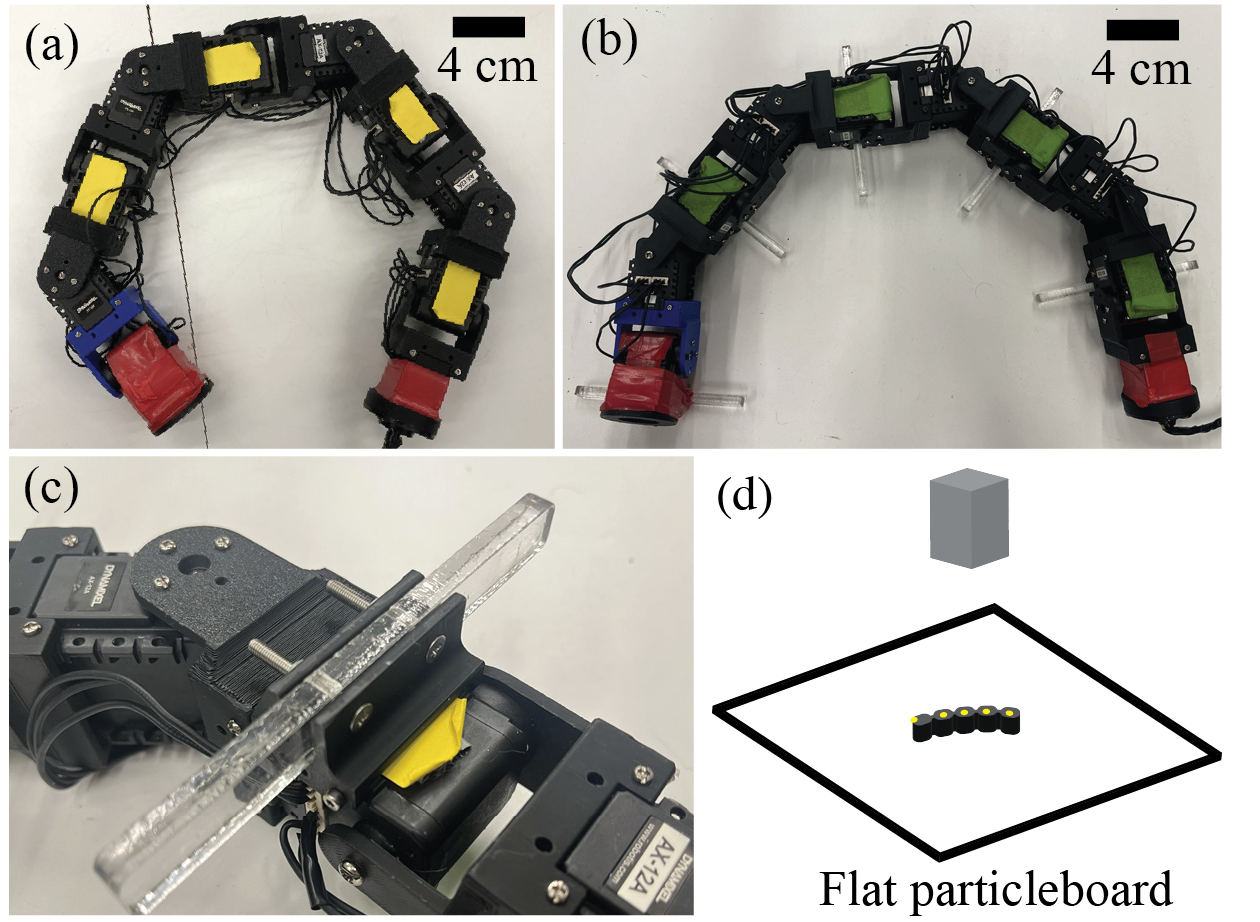}
    \caption{\textbf{The Robophysical Model and Experimental Setup} (a) The robophysical model in the limbless configuration, showing the red markers on the ends for position tracking and yellow markers on the top for orientation tracking. (b) The model in the limbed configuration. (c) The static limbs are attached to the bottom of the model with 3D-printed brackets. (d) The robophysical experimental setup, consisting of a floor of hard-flat particleboard recorded from above with a webcam.}
    \label{fig:fig3}
\end{figure}%

\subsection{Design and Construction}

We constructed a robophysical model (see Fig.~\ref{fig:fig3}a, b) to test our two-wave model by propagating prescribed the lateral and vertical travelling waves. Our model is inspired by the design of modular snake robots \cite{chosetSnake}, consisting of a chain of 9 total servomotors where the axis of rotation for each servomotor is offset 90 degrees from the previous servo's axis of rotation, resulting in a chain of servos providing alternating lateral and vertical degrees of freedom (see Fig.~\ref{fig:fig3}a). This setup allows for the implementation of the two travelling waves prescribed by our model. The 3D-printed joints between servos are labelled by \textit{i} from 1 to 9, where joints with odd \textit{i} are vertical and joints with even \textit{i} are lateral. 

The self-righting strategy we propose is driven exclusively by body undulation in the lateral and vertical planes. However, the presence of legs can introduce additional resistance to rolling~\cite{khazoom2022humanoid} and affect the self-righting effectiveness. To test the effect of legs, we compare two robot morphologies: with and without the legs. Specifically, we prepared static (non-actuated) limbs, made of laser-cut acrylic  attached to the robophysical body via 3D printed brackets (see Fig.~\ref{fig:fig4}c).

Red markers were wrapped around each end of the model for position tracking, and green markers were placed on the top and yellow on the bottom for orientation tracking. The robot was powered by a constant voltage  of 11.3 V from a DC power supply. 

\subsection{Conducting Experiments}

To test the effectiveness of our proposed self-righting scheme, we placed the legged robophysical model on flat, hard ground and tested the one-shot gait. Specifically, we used the gait equation in Eq.~1 and choose amplitude $A=\pi/4$. We ran the gait for a half cycle ($\omega t \in [0,\ \pi])$, and we observed that the robot successfully rolled over upside down (snapshot presented in Fig.~5b.i).

We then gradually decreased the amplitude and investigated whether a minimum amplitude is required to facilitate effective self-righting. We observed that with $A=\pi/12$, the rolling moment generated by body undulation is not sufficient to overcome to energy barrier created by the legs. Thus, the legged robot cannot perform self-righting at lower amplitudes (snapshots presented in Fig.~5b.ii).

Finally, we tested the feasibility of the sequential self-righting strategy. We used Eq.~2 to prescribe the robot gait, choosing $\xi=0.6$ and $A=\pi/4$. We noticed that the robot starts the rolling from the head module, and such rolling behavior propagates from head to tail. At the the end of the gait ($\omega t \in [0,\ \pi]$), the entire robot had experienced successful self-righting (snapshot presented in Fig.~5b.iii).

\begin{figure}
    \centering
    \includegraphics[width=\linewidth]{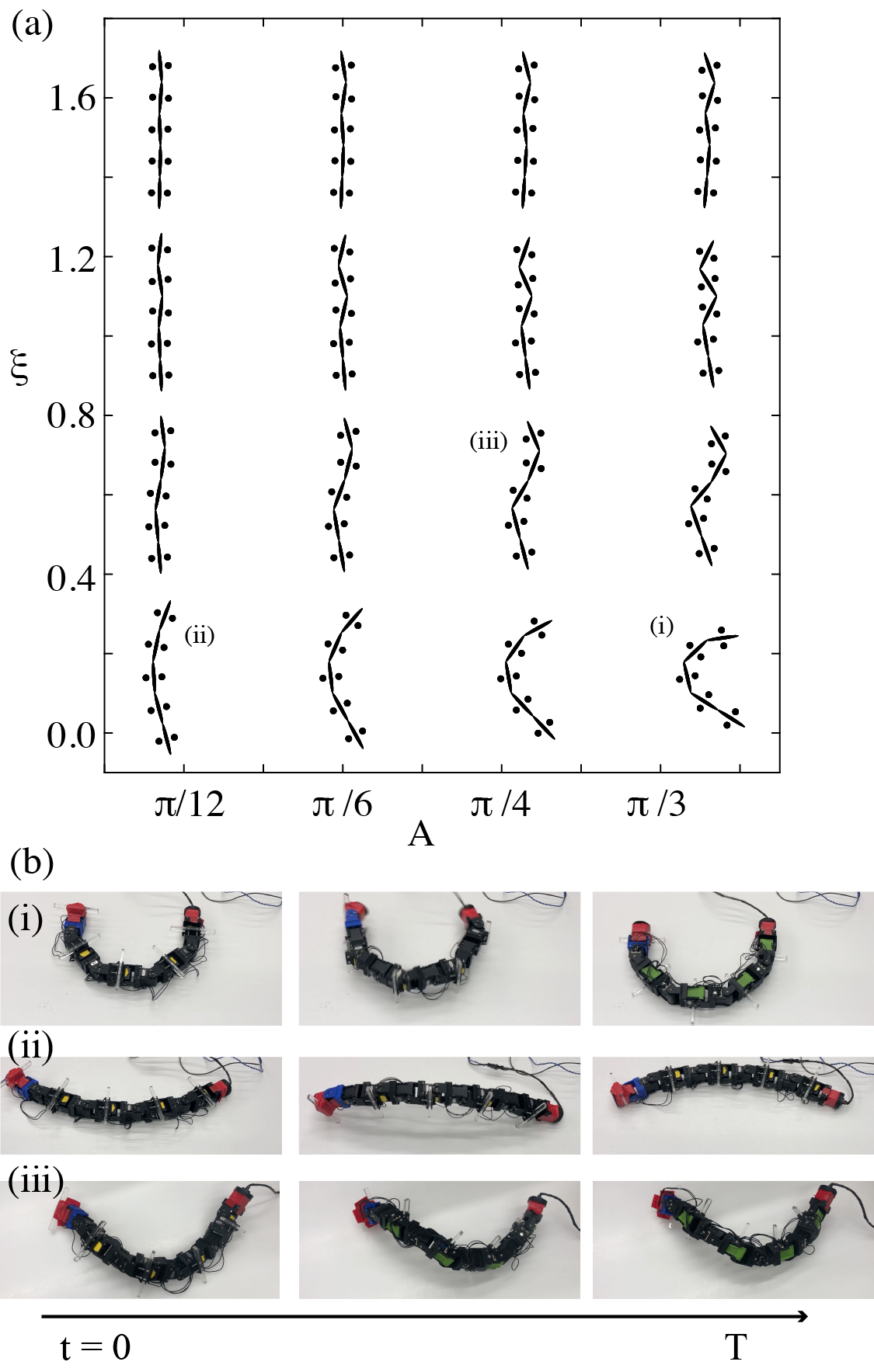}
    \caption{\textbf{Limbed Self-righting} (a) A behavior diagram characterized by (horizontal axis) the body amplitude and (vertical axis) the spatial frequency. We illustrate the top-view postures of the robophysical model over the behavior diagram. (b)(i) Successful one-shot self-righting at a body amplitude of \(\pi/4\). (b)(ii) Unsuccessful one-shot self-righting at a body amplitude of \(\pi/12\), indicating the need of sufficient amplitude for success. (b)(iii) Successful sequential self-righting at a body amplitude of \(\pi/4\).}
    \label{fig:fig5}
\end{figure}

Based on our observation, we argue that the amplitude, $A$, and the spatial frequency, $\xi$, are the key parameters that affect self-righting. Thus, we construct a behavior diagram (Fig.~5.A) of the spatial frequency and the amplitude, and empirically evaluate the self-righting effectiveness over the behavior diagram. The postures of the robot from the top-view over the behavior diagram are illustrated in Fig.~\ref{fig:fig5}a. 

\subsection{Behavior Diagram}

To systematically measure the self-righting effectiveness, we used a top-view camera to record the robot behaviors across the diagram. For each point on the behavior diagram, we ran the gait over three cycles ($\omega t \in [0, 6\pi]$). Each experiment was repeated for 5 trials. If the self-righting is successful, we expect the robot to finish a complete roll ($\Delta \gamma  = 2\pi$, including a self-righting and a self-wronging) over a cycle.

Let $P_{sr}$ be the probability of a self-righting. Then the expected average body rolling displacement, $\Delta \gamma$ per cycle, is $2\pi P_{sr}$. We estimate the probability of successful self-righting by empirically measuring the average rolls per cycle. The number of rolls at each parameter combination relates to self-righting success in the following way:

 \begin{itemize}
     \item $\Delta\gamma = 2\pi$  indicates $P_{sr} = 1$
     \item $\Delta\gamma = 0$ indicates $P_{sr} = 0$
     \item $\Delta\gamma \in (0,\ 2\pi) $ indicate $P_{sr}\in(0,\ 1)$
 \end{itemize}

We swept the behavior diagram with our model for no legs and 11 cm legs, recording five trials at each combination. We used the resolution of $\pi/24$ for $A$, and $0.1$ for $\xi$.

\begin{figure}[t]
    \centering
    \includegraphics[width=0.9\linewidth]{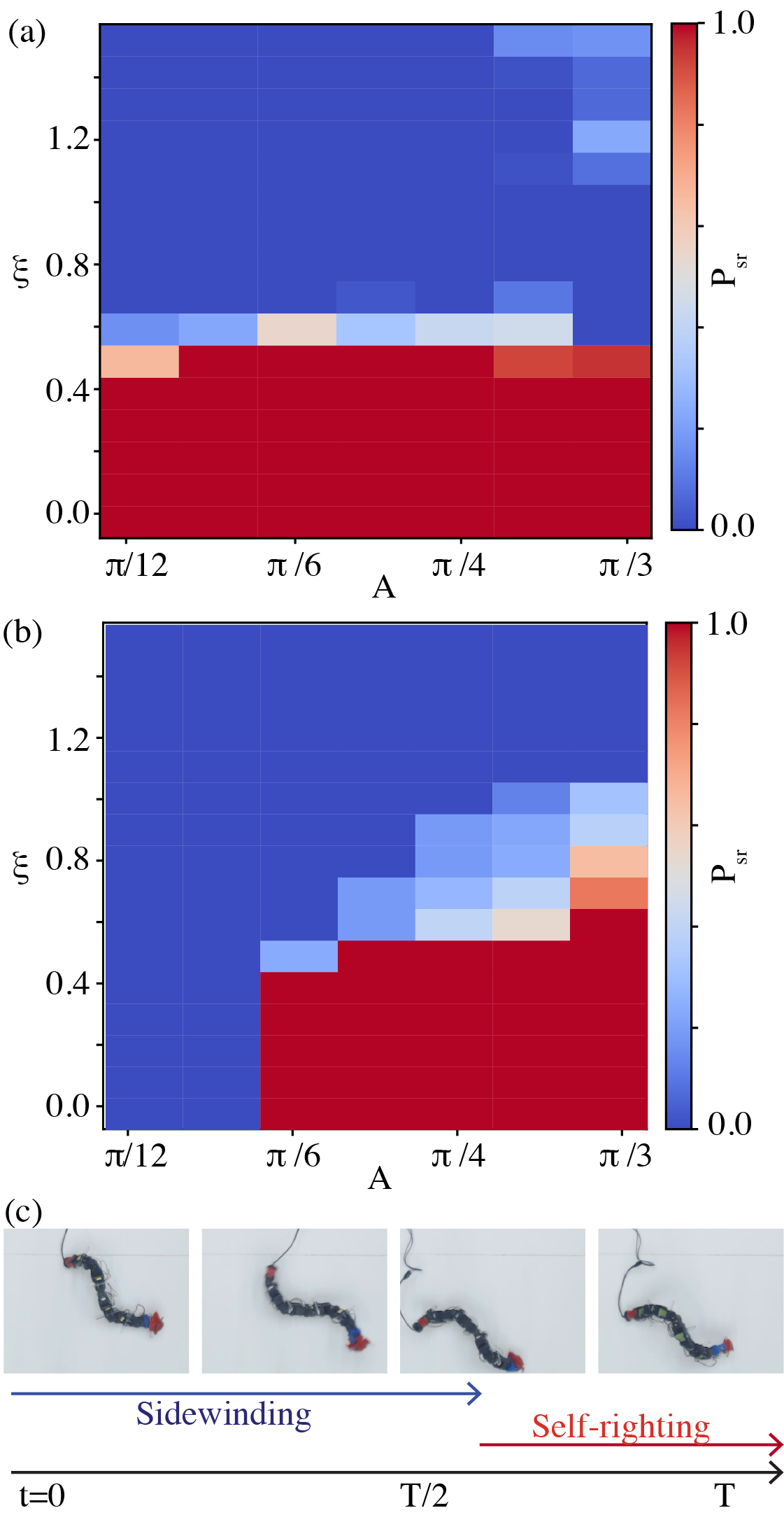}
    \caption{\textbf{Behavior diagram of Self-righting Success} A behavior diagram of self-righting for (a) a multi-legged robot and (b) a limbless robot. Color represents the probability of a successful self-righting ($P_{sr}$). (c) An illustration of intermediate $P_{sr}\in (0,\ 1)$. The robot failed to perform self-righting and result in sidewinding for the first half of a cycle (t = 0 to t = T/2). The robot successfully performs self-righting in the second half of a cycle (t = T/2 to T). The video can be found in SI.} 
    \label{fig:fig6}
\end{figure}%

\begin{figure}
    \centering
    \includegraphics[width=\linewidth]{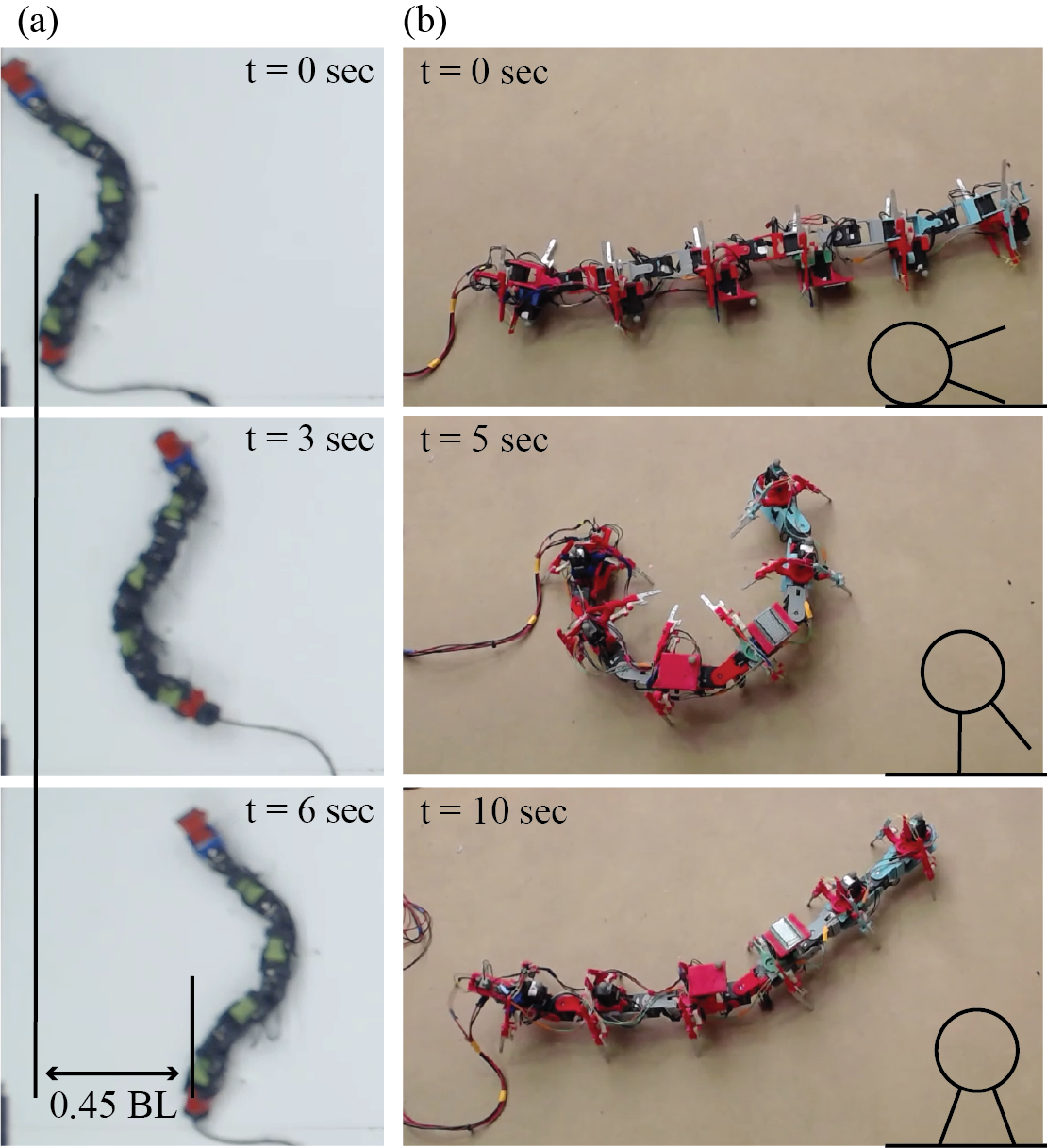}
    \caption{\textbf{Generalized self-righting and sidewinding behaviors} (a) The presence of static limbs enhanced the model's ability to sidewind by increasing gait stability, leading to a lateral displacement of 0.45 body lengths per cycle. (b) A tipped-over multi-legged centipede robot~\cite{chongScience} successfully executes the one-shot self-righting gait.} 
    \label{fig:fig7}
\end{figure}%

\subsection{Results}

Behavior diagrams of the number of rolls per cycle for limbed and limbless configurations are shown in Fig.~\ref{fig:fig6}. We notice that for a limbless robot, the spatial frequency, $\xi$, is the dominant factor that characterizes self-righting effectiveness. For $\xi < 0.5$, self-righting will always be successful even at low amplitudes (e.g., $A=\pi/12$). Moreover, the self-righting effectiveness is almost binary: will have $P_{sr}$ either 0 or 1. We also observed that unsuccessful self-righting resulted in sidewinding.

However, with legs, both $A$ and $\xi$ contribute to self-righting effectiveness (Fig.~\ref{fig:fig6}.a). The boundary between successful and unsuccessful self-righting substantially changes. Specifically, at least $A=\pi/6$ is required for effective self-righting. Further, self-righting with $\xi>0.5$ is feasible with higher amplitude, a feature we did not observe for limbless robots. Finally, we observe a wide margin in the behavior diagram between regions with $P_{sr} = 1$ and $P_{sr} = 0$, indicating a transition zone between successful and unsuccessful self-righting where both behaviors are observed that varies from trial to trial (e.g., Fig.~\ref{fig:fig6}c). Other factors, such as initial condition and inertia, can also contribute to self-righting effectiveness.

\subsection{Self-righting in a more complicated robot}
To ensure that our model is general and can extend to other elongate multi-legged robots, we tested it on three other fully-featured elongate multi-legged robots~\cite{chongScience}. These robots successfully executed both the one-shot (see Fig.~\ref{fig:fig7}b for one example) and sequential strategies, indicating the model's applicability to more elongated limbed robots capable of forward locomotion.

\section{SIDEWINDING}

Our robophysical experiments reveal a spectrum between sidewinding and self-righting in our limbed model. We noticed that in the intermediate regimes and where $P_{sr} = 0$,  complex behaviors can result in a net translation in the lateral direction. This phenomenon resembled the sidewinding observed in limbless organisms such as sidewinder rattlesnakes. We next study the relationship between successful sidewinding and  unsuccessful self-righting. 

Prior work~\cite{chong2022general,chong2021frequency} has documented that rolling in sidewinding gaits can introduce unwanted/unstable body configurations which substantially decrease the robot's sidewinding performance (less distance traveled per cycle). To resist rolling and sidewind effectively, we needed a minimal spatial frequency of $\xi=1.2$, but while this control strategy effectively decreases rolling, it will substantially decrease the sidewinding speed (0.2 Body Lengths per cycle (BL/cyc) for a similar robot as reported in~\cite{chong2023optimizing,chong2021moving}). 

In Sec. IV, we show that the legs can introduce energy barriers to rolling and can cause unsuccessful self-righting. We hypothesize that the presence of legs can also stabilize the sidewinding gaits and thus enable effective sidewinding at lower spatial frequency $\xi$ (and thus better sidewinding performance).

To test our hypothesis, we ran our robot with a vertical amplitude of \(\pi/9\)  and a lateral amplitude of \(\pi/3\), and a spatial frequency of $\xi=0.6$. We observed a lateral displacement of 0.45 body lengths per cycle (see Fig.~\ref{fig:fig7}a), substantially greater than has been achieved with limbless robots of similar length. This indicates that limbed sidewinding shows potential for more stable locomotion than limbless sidewinding.

\section{CONCLUSION AND FUTURE WORK}

Here we developed an effective and general self-righting scheme for elongate limbed systems which substantially increases the robustness of these robots, which are particularly susceptible to falling over. We based our schemes on  preliminary centipede self-righting experiments, though more work remains to be done on quantifying this animal behavior. Our model can now be applied to elongate robots with orientation sensing so they can self-right automatically after detecting a tip-over. We have also explored the viability of using changes in body shape, as opposed to active limbs or passive rolling, to self-right successfully. This method has an advantage over limb-based methods in elongated robots because it does not require the addition of extra limbs or limb range of motion, but relies only on changes in body shape as a result of degrees of freedom already present.

In future work, we will further evaluate the behavior diagram of self-righting by considering other metrics, such as efficiency and robustness to perturbation. Li et al. have documented the potential energy landscape for different strategies of self-righting in cockroaches, characterizing the potential energy barrier that needs to be overcome for the organism to self-right using different strategies \cite{li2019cockroach}. This approach could be applied to the two strategies we have characterized, informing which method a robot in the real world should select if tipped over. Additionally, our study  yields further insight into biological self-righting, demonstrating that centipedes  exhibit multiple self-righting behaviors that rely on body shape changes, rather than limb use.

\section*{ACKNOWLEDGMENTS}

The authors would like to thank Esteban Flores for his help running experiments and his discussion and Chris Pierce for his image analysis expertise. They would also like to thank the Georgia Tech Physics REU and the National Science Foundation for their logistical, educational, and financial support (NSF Grant 2244423). The authors would also like to thank the Institute for Robotics and Intelligent Machines at Georgia Tech for the use of experimental space. The authors received funding from NSF-Simons Southeast Center for Mathematics and Biology (Simons Foundation SFARI 594594), the Army Research Office grant W911NF-11-1-0514, and a Dunn Family Professorship. 

\clearpage
\addtolength{\textheight}{-12cm}   





\bibliographystyle{unsrt}
\bibliography{bib}

\end{document}